\documentclass[conference, 11pt]{IEEEtran}
\IEEEoverridecommandlockouts
\usepackage{amsmath,amssymb,amsfonts}
\usepackage{algorithmic}
\usepackage{graphicx}
\usepackage{textcomp}
\usepackage{xcolor}
\usepackage{soul}
\usepackage{longtable}
\usepackage{subcaption}
\usepackage{url}

\renewcommand{\hl}[1]{#1}

\definecolor{lightgreen}{rgb}{0.8, 1, 0.8}
\definecolor{lightred}{rgb}{1, 0.8, 0.8}

\def\BibTeX{{\rm B\kern-.05em{\sc i\kern-.025em b}\kern-.08em
    T\kern-.1667em\lower.7ex\hbox{E}\kern-.125emX}}
    
\begin{document}

\title{The Distribution Shift Problem in Transportation Networks using Reinforcement Learning and AI \thanks{Paper accepted for publication by IEEE Transactions on Intelligent Transportation Systems}}

\DeclareRobustCommand*{\IEEEauthorrefmark}[1]{%
  \raisebox{0pt}[0pt][0pt]{\textsuperscript{\footnotesize\ensuremath{#1}}}}

\author{
    \IEEEauthorblockN{
    F. Taschin \IEEEauthorrefmark{2, 3},
    A. Lazraq \IEEEauthorrefmark{3},
    I. Ozgunes \IEEEauthorrefmark{3},
    O. K. Tonguz \IEEEauthorrefmark{1, 3}}
    
    \\

    \IEEEauthorblockA{\IEEEauthorrefmark{1}Carnegie Mellon University}
    \IEEEauthorblockA{\IEEEauthorrefmark{2}KTH Royal Institute of Technology}
    \IEEEauthorblockA{\IEEEauthorrefmark{3}Virtual Traffic Lights LLC}
}

\maketitle

\begin{abstract}
The use of Machine Learning (ML) and Artificial
Intelligence (AI) in smart transportation networks has increased significantly in the last few years. Among these ML and AI approaches, Reinforcement Learning (RL) has been shown to be a very promising approach by several authors.
However, a problem with using Reinforcement Learning in Traffic Signal Control is the reliability of the trained RL agents due to the dynamically changing distribution of the input data with respect to the distribution of the data used for training. This presents a major challenge and a reliability problem for the trained network of AI agents and could have very undesirable and even detrimental consequences if a suitable solution is not found. Several researchers have tried to address this problem using different approaches. 

In particular, Meta Reinforcement Learning (Meta RL) promises to be an effective solution.  In this paper, we evaluate and analyze a state-of-the-art Meta RL approach called MetaLight and show that, while under certain conditions MetaLight can indeed lead to reasonably good results, under some other conditions it might not perform well (with errors of up to 22\%), suggesting that Meta RL schemes are often not robust enough and can even pose major reliability problems. 
\end{abstract}


\section{Introduction}

As cities become more populated and the number of vehicles on their roads increases, the problem of efficiently controlling the flow of vehicles to reduce travel times and CO2 emissions is becoming a top priority. For this reason, in recent years, research in Traffic Signal Control has gained significant momentum as the quest to develop better Traffic Signal Control algorithms intensified. Advancements in hardware, such as cheaper and more powerful radar systems, Lidars, and cameras, as well as in software for Machine Learning and Deep Learning, sparked the development of AI-powered Traffic Signal Control algorithms. Specifically, Deep Reinforcement Learning (Deep RL) gained much attention in the research community as it better captures the sequential decision-making nature of the problem. 

Deep Reinforcement Learning algorithms can learn optimal Traffic Signal Control policies by interacting with their environment (a realistic simulator of traffic dynamics) in a trial-and-error fashion, \hl{training a Neural Network to make traffic signal control decisions in real-time}. In contrast to rule-based algorithms, such as MaxPressure \cite{maxpressure} which aims to minimize congestion by dynamically adjusting signal timings based on the principle of \hl{minimizing the pressure, or occupancy, of approaching lanes}, Deep Reinforcement Learning algorithms can generally exploit the observed traffic patterns to achieve better performance. Among these, PressLight \cite{presslight} showed exceptional performance by leveraging a pressure-based reward. FRAP \cite{frap}, on the other hand, introduced an invariant model capable of operating on intersections of different shapes. Multi-Agent Deep Reinforcement Learning algorithms can further improve the performance by establishing coordination between signalized intersections. CoLight \cite{b7}, for example, achieves global communication and coordination through the application of Graph Attention Networks.

Adoption of these algorithms, however, has not been forthcoming. A main obstacle to deploying Deep Traffic Signal Control algorithms in the real world is \textit{Distribution Shift}. \hl{Distribution shift occurs when a model trained on data with a particular probability distribution is evaluated on data drawn from a different distribution. This can happen, for example, when an image classification model encounters objects it wasn't trained on, or when a traffic signal control algorithm faces atypical traffic conditions—such as those caused by a major concert or a severe accident}. Deep Learning in general, and Deep Reinforcement Learning in particular, greatly suffer when the input distribution at deployment time (or in the real world) significantly shifts from the distribution used during the training time. As an example, if we were to train a Deep Reinforcement Learning model to control the traffic signals at an intersection where most of the traffic flows in the North-South direction, its performance would drop when deployed in an intersection where the busiest direction is East-West, causing longer queues and congestion. 

To deal with the Distribution Shift problem, MetaLight \cite{metalight} proposes to train a Deep Reinforcement Learning model to \textit{adapt} to a new scenario (where traffic distribution or intersection shape changes) given some examples, \hl{i.e. recordings of traffic state and traffic signal control decisions over time. Here, instead of training a model to optimally control traffic for a specific set of scenarios, MetaLight trains a neural network model on a wide array of scenarios with a specific objective function and training mechanism so that it can be adapted to new ones with a small amount of additional data}. 
This important work shifts the key paradigm from \textit{training a model able to generalize to any scenario} to \textit{training a base model that can be adapted to any scenario}. \hl{This paradigm is somewhat similar or synergistic to developing a \textit{foundational model} for traffic signal control, where a base  model is trained on a large amount of data to be able to operate in diverse situations. In this work, we analyze MetaLight (proposed by }\cite{metalight}\hl{)} \hl{performance for increasing levels of distribution shift, measuring it as described in} \cite{tonguz2025}.

The remainder of this paper is organized as follows. In Section II, we summarize the main contributions of this paper. Section III summarizes the related works. While Section IV describes the Distribution Shift Problem, Section V explains how Reinforcement Learning is used for smart Traffic Signal Control, and Section VI discusses how distribution shift affects it. Section VII summarizes the MetaLight framework used in the experiments, described in Section VII (Experimental Setting) and Section IX (Experiments). Finally, in Section X we present the results, and in Sections XI (Discussion) and XII (Conclusions), we analyze them and discuss their implications in real-world smart traffic signal control.

\section{Contribution}
In this paper, we contribute to the ongoing study of the Distribution Shift issue in the area of Reinforcement Learning for  Traffic Signal Control with a systematic analysis and evaluation of an existing state-of-the-art algorithm, MetaLight, that aims to solve the Distributional Shift issue by generating Deep RL models that can be adapted to new traffic distributions. We evaluate the algorithm in a wider variety of scenarios than the original paper, using both synthetic and real-world data. We analyze MetaLight performance degradation as a function of the distance of the test scenario distribution from the training distribution using the Kullback-Leibler Divergence as a statistical distance metric, \hl{extending the (relatively little) literature on the measurement of distribution shift by providing new measurements and relative results for an important algorithm in the field (MetaLight). Moreover, we evaluate its performance using scenarios simulated from real-world data and road networks.}

\section{Related Work}
The Distribution Shift problem has been studied since before the advent of Neural Networks. Many excellent works provided theoretical frameworks to frame the issue, with \cite{pac} being arguably the most influential. Many works highlighted the problem over the years, from non-Deep Learning models as in  \cite{concept_drift}, \cite{concept_drift_survey}, and \cite{correcting} to the famous Domain-Adaptive Network architecture \cite{domain_adaptive} in deep object recognition, and unsupervised methods \cite{unsupervised_adaptation}. Interestingly, even the now-famous BatchNorm layer \cite{batchnorm}—a neural network layer extensively used in state-of-the-art architectures—was developed to mitigate the distributional shift problem.

In recent years, the problem of optimizing Traffic Signal Control through Reinforcement Learning has gained a lot of attention. Rule-based Traffic Signal Control algorithms, such as MaxPressure \cite{maxpressure} and SOTL \cite{sotl} achieve higher performance than pre-timed and fixed-cycle traffic signal control policies, but cannot exploit additional knowledge of traffic patterns. Therefore, Reinforcement Learning promises to bring additional benefits as RL agents can learn the optimal policy for every scenario. Authors of \cite{actorcritic} showed that high-performance Traffic Signal Control could be achieved with Actor-Critic Reinforcement Learning methods. \cite{2016rl} is one of the earliest works to use the now popular DQN RL algorithm \cite{mnih2013playing} in Traffic Signal Control with a handcrafted reward function. PressLight \cite{presslight} achieved high performance by designing a reward function based on the MaxPressure algorithm. Further work \cite{thousandlights} scaled the PressLight approach to a large number of intersections, showing that a pressure-based reward can function well in a decentralized manner and a wide range of scenarios. However, the pressure reward is designed for networks where traffic lights are all relatively close to each other, which is not always the case. \cite{frap} introduces a novel Neural Network architecture that can be used on a wide range of intersection shape distributions, e.g., 3-way intersections or intersections where not all traffic movements are available. \cite{b7} instead proposes a Graph-Attentional Neural Network architecture to promote coordination between intersections and reduce travel times on global scale. 

Although some of the aforementioned works approach the performance limits of Traffic Signal Control in some scenarios, they do not address the Distribution Shift problem, as they don't focus on the cases where the traffic distribution at test time is very different from the distribution at the training time. In addressing the Distributional Shift problem in Traffic Signal Control, MetaLight \cite{metalight} is one of the most important works, proposing a Meta-RL algorithm to train models that can be adapted to new scenarios with less data and training time than it would take to re-train them from scratch. GeneraLight \cite{generalight} improves on it by developing a method to simulate a wider range of realistic traffic distributions on which to train MetaLight. Although this seems to improve adaptation performance, it adds a lot of complexity and does not change the fundamental way MetaLight operates, therefore we do not use this technique in our analysis. A similar work that also exploits Meta-Learning is \cite{meta_gnn}, but it also employs Graph Attention Networks to allow for multi-intersection control. MetaSignal \cite{metasignal} also uses Meta-Learning in combination with a Fourier basis function representation of the value function, which reduces instability and allows for better adaptation at the expense of less expressive RL policies.

In a parallel line of work, \cite{datamightbeenough} shows excellent performance in simulation-to-real-world adaptation, which often is a much more difficult problem in terms of Distribution Shift. However, it relies on the DynamicLight algorithm \cite{dynamiclight}, where the RL agent only determines the duration of the greedily chosen action. This results in agents more robust to Distributional Shift, at the expense of less expressive policies.

\hl{The problem of how to measure distribution shift in RL-based control of traffic scenarios is under-studied. To the best of our knowledge,} \cite{tonguz2025}  \hl{is the only work that specifically addresses this problem, using the KL divergence measure of the vehicle volume distribution as a proxy to measure the difference between traffic scenarios. Generalight} \cite{generalight} \hl{also measures scenario-to-scenario distance, but considers each vehicle route as a separate object, neglecting therefore the fact that since different routes can share road segments, two traffic scenarios could be very similar even if their individual routes differ.}

\section{The Distribution Shift Problem}
\subsection{Notation}
\hl{Assume we have a neural network that we want to optimize to perform a certain task} $T$ ($T$ \hl{can be a classification task, a regression, or a control task such as traffic signal control).} 
Let $X$ denote the input space of the task $T$ and $Y$ denote the output space. \hl{For example, in traffic signal control,} $X$ \hl{could be the queue lengths at an intersection and} $Y$ \hl{an action that tells which approach to serve a green light to.}
$P_{train}(X,Y)$ represents the joint distribution of inputs and outputs in the training (source) dataset and $P_{test}(X,Y)$ represents the joint distribution of inputs and outputs in the test (target) dataset. It is almost always the case that $P_{train} \neq P_{test}$; i.e., the distribution of data used for training is not the same as the one encountered when testing the model in the real world.

\subsection{Distributional Shift}
A neural network model can be represented as a function $f:X\rightarrow Y$ that maps inputs to outputs. The goal of the training process is to find the function $f$ that minimizes the expected loss over the training distribution, typically represented by a loss function $L:Y\times Y \rightarrow \mathbb{R}$, e.g., the mean squared loss between the neural network outputs and the actual labels. In Deep Neural Networks, $f$ is parameterized by a set of weights and biases that we call $\theta$. The objective of the training is to find the optimal value of $\theta$ that minimizes the expected loss over the training distribution.

\begin{equation}
\label{eq:train}
    \theta_{train}* = \arg\min \mathbb{E}_{(x, y) \sim P_{train}}[\mathcal{L}(f_{\theta}(x), y)]
\end{equation}

However, in the presence of a distributional shift, the performance of the model is evaluated on the test distribution, which leads to the true objective we care about in deployment:

\begin{equation}
    \theta_ {train}* = \arg\min \mathbb{E}_{(x, y) \sim P_{test}}[\mathcal{L}(f_{\theta}(x), y)]
\end{equation}

Generally, $\theta_{train}$ is different from  $\theta_{test}$, which leads to the Distributional Shift problem: the model obtained during training by optimization of Equation \ref{eq:train} is not the optimal model for the test distribution. 

\subsection{Types of Distributional Shift}
\begin{figure}
    \centering
    \includegraphics[width=0.45\textwidth]{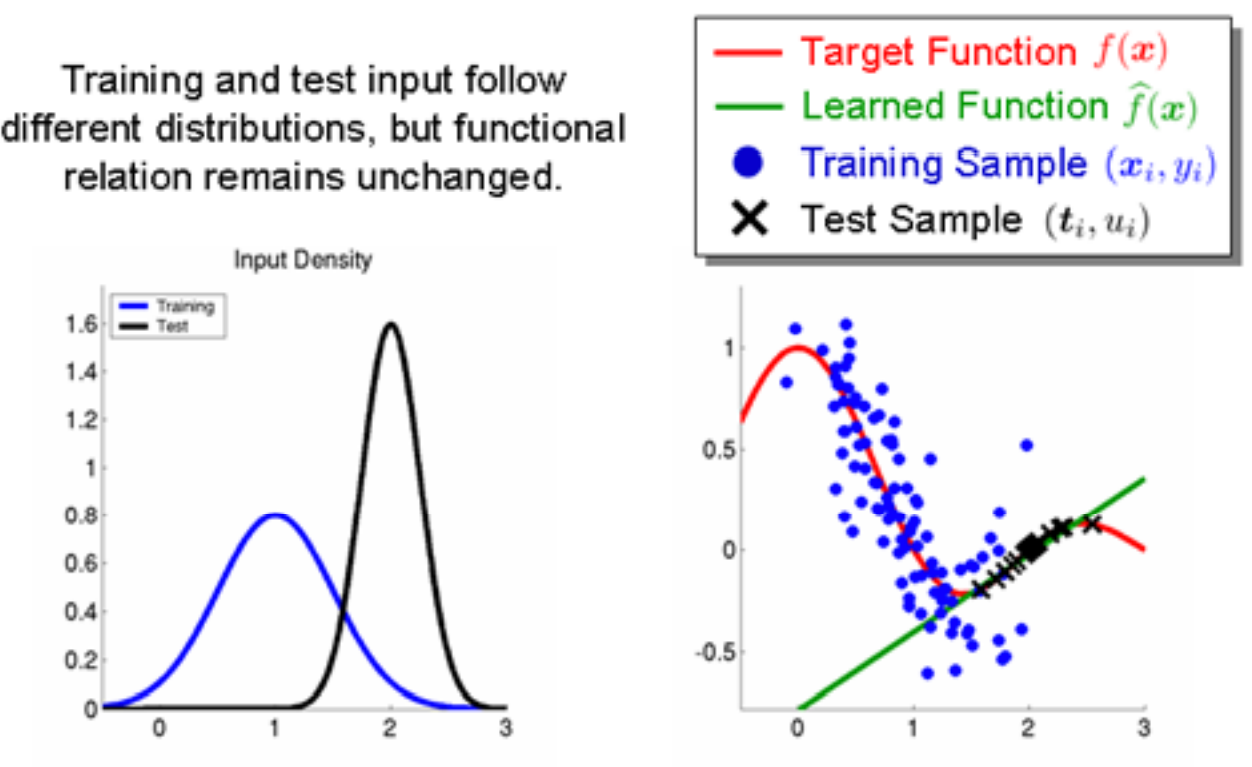}
    \caption{Covariate Shift after \cite{cov_shift_post} : training data does not cover the portion of data observed at test time, therefore the learned model performance greatly decreases at test time.}
    \label{fig:covariate}
\end{figure}

Distributional Shift can be categorized into three main types:
\paragraph{Covariate Shift}: Where the input distribution changes, but the conditional distribution of the output given the input remains the same: $P_{train}(X) \neq P_{test}(X)$ but $P_{train}(Y|X) == P_{test}(Y|X)$. The problem here is just a lack of data points in the training set, and, when possible, can be solved by adding more training data. Figure \ref{fig:covariate} shows a simplified example of Covariate Shift: the train set contains data points from only a portion of the whole distribution, and a model fit to those points cannot predict correctly for data points in the test set. In the case of Traffic Signal Control, where the Neural Network input consists of the number of vehicles in each lane, a Covariate Shift could happen when the traffic volume changes and the distribution of vehicles in each lane drifts from that observed during training.

\paragraph{Label Shift}: is the same as Covariate shift, but for the labels $Y$; i.e., $P_{train}(Y) \neq P_{test}(Y)$ but $P_{train}(Y|X) == P_{test}(Y|X)$.

\paragraph{Concept Shift}: Happens when the relationship between inputs and outputs changes between train and test time, i.e., $P_{train}(Y|X) \neq P_{test}(Y|X)$. In Supervised tasks, this might be due to the partial-observable nature of the task (e.g., in a movie recommender system, the personal preferences of a user might not be directly observable) or the non-stationarity of the environment (some genres can lose interest over time as it happened to Western movies). In the context of Traffic Signal Control, when the traffic distribution changes, for the same instantaneous observation $X$ could correspond to an evolution of the traffic at the intersection over the next minutes that is very different from how the traffic evolved in the training distribution if starting from the same observation $X$.

\begin{figure}
    \centering
    \includegraphics[width=0.45\textwidth]{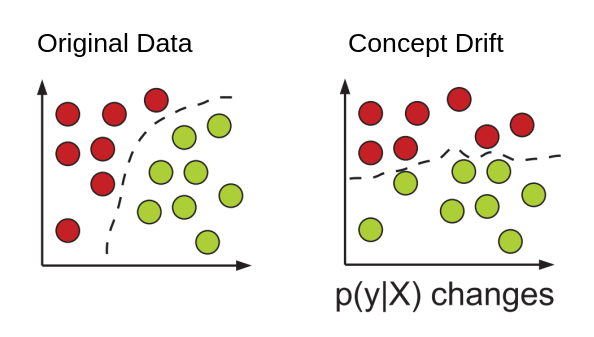}
    \caption{Concept Drift after \cite{concept_drift_survey}: the distribution of data points does not change while $P(y|x)$ does. In this example, $y$ is the color of a point given its $x$ coordinates.}
    \label{fig:enter-label}
\end{figure}

\section{Reinforcement Learning for Traffic Signal Control}
\label{sec:rl}
\begin{figure}
    \centering
    \includegraphics[width=0.45\textwidth]{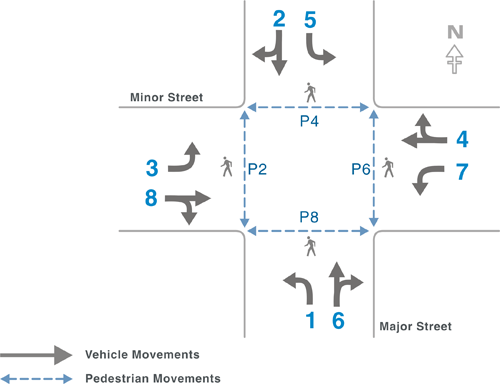}
    \caption{Possible traffic movements at a signalized intersection}
    \label{fig:traffic_movements}
\end{figure}

The Traffic Signal Control problem is framed in Reinforcement Learning as a Markov Decision Process (MDP) with state space $S$, action space $A$, and a reward function $r(s, a) \rightarrow \mathbb{R}$. The state is an observation of the traffic at the intersection (e.g., queue lengths) and an action is a decision on which \textit{phase} to serve. A phase is a combination of traffic movements (see Figure \ref{fig:traffic_movements}) that can be served (i.e., receive a green light) at the same time. Since right turns are controlled together with through movements, we do not count them as separate movements, and the terms "phase" and "movement" can be used interchangeably in this paper. Transitions between states are stochastic and are regulated by the -unknown- transition probability function $P(s' | s, a)$ that relates the probability of a new state $s'$ given the current state $s$ and the chosen action $a$.

The goal is to discover an optimal policy $\pi: S \rightarrow A$ that maps the observed state of the intersection to an action such that the cumulative discounted reward is maximized. In mathematical terms, the cumulative discounted reward at time $t$, that we will call from now on the \textit{return} is defined as
\begin{align}
    R_t = \mathbb{E}\left[\sum_{k=t}^{\infty} \gamma^{t-k} r_k \right]
\end{align}
which is the expected value of the sum of future rewards, discounted by a discount factor $\gamma$.

In value-based Deep Reinforcement Learning, a Neural Network is trained to learn a function that maps each state-action pair to the optimal return
\begin{align}
    \label{eq:optimal_q}
    Q^*(s, a) = \max_{\pi}\mathbb{E}_{\pi}\left[\sum_{k=t}^{\infty} \gamma^{t-k} r_k | s_t = s, a_t = a\right]
\end{align}
which is, the expected return of the optimal policy $\pi^*$ (i.e., the policy that maximizes the expectation) that, starting from state $s$ and action $a$, continues to choose optimal actions such that the quantity $R$ is maximized. The Neural Network is updated with Gradient Descent to minimize the Bellman Squared Loss
\begin{align}
\label{eq:bellman_loss}
    \mathcal{L}(s, a, r, s') = \left(Q(s, a) - r(s, a) - \gamma \max_{a'}Q(s', a')\right)^2
\end{align}
where $s'$ is the state we end up in by choosing action $a$ in state $s$. Repeatedly minimizing this loss from transitions $(s, a, r, s')$ experienced by interacting with the environment generally produces a Neural Network that estimates with good approximation the optimal $Q^*$ function and allows us to choose optimal actions $a^*$ that maximize $Q(s, a*)$ in any state $s$.

\section{Distribution Shift in Traffic Signal Control}
As discussed in Section \ref{sec:rl} above, one can frame the Traffic Signal Control problem as a Markov Decision Process governed by the transition probability function $P(s'| s, a)$. This function, which is not known a priori, incorporates all the stochastic dynamics of real-world traffic, such as drivers' behavior, the arrival rate of vehicles, vehicles' speed and acceleration, etc. If $s$ is the number of vehicles in each road of the intersection, $P(s'|s, a)$ would be greatly influenced by the future arrival rate of vehicles, which can vary for many unobservable reasons. In the real world, the probability $P(s' | s, a)$ is generally \textit{non-stationary}; i.e., it changes over time. While some of its variations can be predicted (e.g., by knowing the expected vehicular arrival rate for every hour and day of the week), many unobservable or unpredictable factors make it hard to predict such variations (e.g., car accidents, anomalous traffic due to particular events, etc). This is particularly true when deploying an RL agent on a new intersection that has significantly different dynamics than those observed during training. As $P(s'|s, a)$ varies, the optimal Q values of Equation \ref{eq:optimal_q} become obsolete, and the performance of the RL agent degrades.

\section{MetaLight Framework Overview}
\label{sec:metalight}

The MetaLight framework \cite{metalight}, introduced by Xinshi Zang et al. in 2020, addresses the Distributional Shift problem in traffic signal control by proposing a novel approach that leverages meta-reinforcement learning techniques. The primary goal is to enhance the learning efficiency of Deep Reinforcement Learning models in new scenarios by leveraging the knowledge learned from existing scenarios.

\subsection{Model Improvements: FRAP and FRAP++}

The MetaLight framework builds upon the structure-agnostic and parameter-sharing RL model called FRAP, as introduced by Zheng et al \cite{frap}. FRAP is a novel reinforcement learning model for traffic signal control that modifies the Neural Network's head to produce a phase competition matrix, i.e., a matrix where each entry corresponds to the value of serving a particular phase pair instead of the others. This results in a model capable of dealing with different intersection shapes, converging faster, and demonstrating superior generalizability across diverse road structures and traffic conditions. To enhance flexibility, MetaLight introduces an improved model, FRAP++, which overcomes the limitations of FRAP in handling different lane combinations. While FRAP considered the sum of each lane demand, which can vary widely across intersections due to the different number of lanes and traffic volume,  
FRAP++ represents phase demand as the mean of each lane's demand, reducing this variability problem.

\subsection{Meta-Learning}
Meta-Learning is a paradigm in Deep Learning to design and train deep models that can be adapted to new tasks with minimal additional data and training. For example, with Meta-Learning, one can train a base model to be adapted to recognize images from an object class not present in the training data with just a few images and training steps. Some important meta-learning algorithms like SNAIL \cite{mishra2018a}, LSTM Meta-Learners \cite{ravi2017optimization}, and Prototypical Networks \cite{snell2017prototypicalnetworksfewshotlearning} were developed for few-shot learning tasks. SNAIL combines temporal convolutional networks with attention mechanisms for capturing contextual information. LSTM Meta-Learners use Long Short-Term Memory Networks to optimize another learner's parameters. Prototypical Networks employ metric learning, classifying data points based on their proximity to class prototypes. Arguably, the most famous meta-learning algorithm is Model-Agnostic Meta-Learning (MAML), introduced by \cite{maml}. MAML seeks to find a model parameter initialization that is sensitive to fine-tuning on a variety of tasks, thereby facilitating rapid adaptation. Formally, MAML optimizes for a parameter vector $\theta \in \mathbb{R}^{l}$, where $l$ is the parameter space dimension, such that a single gradient update on a new task $\mathcal{T}_i$ with loss $\mathcal{L}_{\mathcal{T}_{i}}$ brings significant improvements in the model performances on task $\mathcal{T}_i$. Given a fine-tuning function $U_{i}: \mathbb{R}^{l} \rightarrow \mathbb{R}^{l}$ that fine-tunes a deep model for the new task $\mathcal{T}_i$, MAML optimizes the average loss 
\begin{equation}
    \min_{\theta}\mathbb{E}[\mathcal{L}_{\mathcal{T}_{i}}(U_{i}(\theta))]
\end{equation}
In simpler terms, MAML optimizes a model parametrized by $\theta$ to achieve high performance after the fine-tuning function is applied. In practice, the fine-tuning function is often a simple Gradient Descent update step. This means that MAML optimizes a base deep model to respond well to an additional gradient step update. Models trained with MAML can therefore be adapted to new tasks with a small amount of data and a handful of gradient steps (sometimes even a single one). On the contrary, a single gradient step for a new task on a traditional model will have almost no effect, or could even lead to a performance decrease in all tasks.  

\subsection{MetaLight}
MetaLight follows a traditional gradient-based Meta Reinforcement Learning approach, inspired by the aforementioned Model-Agnostic Meta-Learning (MAML) \cite{maml}. To the best of our knowledge, MetaLight is the first algorithm to employ MetaLearning on Traffic signal control. MetaLight introduces significant improvements to MAML that are tailored for value-based RL algorithms like DQN \cite{mnih2013playing}. The base MetaLight model is a FRAP++ model trained with the DQN algorithm to predict the optimal Q value for every possible action as described in Equation \ref{eq:optimal_q} by minimizing the Bellman Loss of Equation \ref{eq:bellman_loss}.
In the Meta-Learning context for Traffic Signal Control, each task is a scenario $\mathcal{S}_{i}$ containing a particular intersection and traffic distribution. The desired optimal model is a model that, after being trained on a set of scenarios $S = \{\mathcal{S}_{i}\}$, can be quickly adapted to a new scenario $\mathcal{S}_{k} \notin S$ with minimal data and gradient updates. In practice, this means that the FRAP++ model (which is a Deep model that predicts Q values for each action) can be adapted to produce the correct Q values for the new scenario with a small number of transitions $(s, a, r, s')$ (see Section \ref{sec:rl}) and gradient updates.  

The MetaLight training alternates two phases: individual-level adaptation and global-level adaptation. In the first, the FRAP++ model is trained to minimize the Bellman loss of Equation \ref{eq:bellman_loss} for a scenario $\mathcal{S}_{k}$ At each time step, the parameters $\theta$ are updated at each step $k$ with a batch sampled from memory $\mathcal{D}_i$ as
\begin{equation}
    \label{eq:individual_level}
    \theta_{k+1} = \theta_{k} - \alpha \nabla \mathcal{L}(\theta_{k}, \mathcal{D}_{i})
\end{equation} 
where $\nabla$ is the gradient operator applied on the weights $\theta$ with respect to the loss, and $\alpha$ is the learning rate. This is performed for all training scenarios. For each scenario, we therefore have an initial set of parameters $\theta_{0}$ which are updated to $\theta_{k}, \theta_{k+1}, ...$ with a series of gradient steps. In the global-level adaptation, all these updates are collected and the initial parameters $\theta_0$ are updated from a newly generated batch of experience $\mathcal{D}'_{i}$ from a set of test scenarios $S^{test} = \{\mathcal{S}_{i}^{test}\}$ as
\begin{equation}
    \label{eq:global}
    \theta_0 = \theta_0 - \beta \nabla_{\theta}\sum_{S^{test}_{i}}\mathcal{L}(\theta, \mathcal{D}'_{i})
\end{equation}
where, again, $\nabla$ is the gradient operator applied on the weights $\theta$ with respect to the loss, and $\beta$ is the learning rate. In simpler terms, the initial parameters $\theta_{0}$ are updated with the gradients from the individual adaptation stage in the direction of minimal test loss. This results in a new set of initial parameters that will have a lower loss after some gradient steps are performed on it.

The framework is as follows: 
a batch of tasks is sampled, and during each interval t, the base learner inherits the initialization from the meta-learner and conducts individual-level adaptation using samples drawn from memory at each time step. At the end of each interval t, the meta-learner takes global-level adaptation with another batch of samples from the memory.

%
%

\subsection{Experiments and Results}

Experiments were conducted using the CityFlow \cite{cityflow} simulation platform, providing realistic traffic signal control environments. Four real-world datasets from different cities were utilized, and scenarios were categorized into three tasks: homogeneous, heterogeneous, and scenarios from different cities. MetaLight was compared against several baselines, including Random, Pretrained, MAML, and Self-Organizing Traffic Light Control (SOTL) \cite{sotl}.

Results reported in the MetaLight paper demonstrated the superior performance of MetaLight, especially in heterogeneous and cross-city scenarios. The framework exhibited faster learning speeds, more stable convergence, and significant improvements in travel time metrics compared to baseline methods.

\section{Experimental setting}
In this section, we describe the experimental setting used in our study.

\subsection{Implementation}
In this paper, we evaluate the MetaLight framework in scenarios of varying traffic distribution. We make the distribution of traffic vary both in terms of arrival rate for each traffic movement (phase) and for the total volume of traffic. To simulate such scenarios we use the CityFlow simulator, as used in the MetaLight paper. The MetaLight implementation we use is the one provided by the authors \cite{metalight_github} and we kept the default hyperparameters for the experiment.

\subsection{Setting}
The MetaLight paper compares the model's performance after performing the adaptation step on the base model trained with the MetaLight approach (see Section \ref{sec:metalight}) versus performing the same adaptation step on a FRAP++ model that was pre-trained with the common RL training described in Section \ref{sec:rl}. The paper also provides a comparison with the adaptation step on a random Neural Network initialization and with the rule-based SOTL algorithm.

In our experiments, we compare the MetaLight adaptation to a given scenario with a FRAP++ model that is trained specifically for that scenario. We find this comparison more relevant for real-world deployment of Deep Reinforcement Learning for Traffic Signal Control, as it is often the case that \textbf{to deploy on a new intersection new training "from scratch" is required. }

In our experiments we focus on a single intersection where only the traffic distribution varies, to better understand the intrinsic capability of MetaLight to deal with traffic distribution shift. Adding other elements, like a multi-agent setting or shape-varying intersections could lead to results that are difficult to interpret.   

\subsection{Metrics}
\label{sec:metrics}
\paragraph{Performance Metric (Travel Time)} We compare algorithms by looking at the Travel Time metric as in the MetaLight paper. Depending on the reward function, looking at travel time alone could be misleading as the agent might learn to sacrifice some throughput to reduce travel times. However, FRAP (the base model used in MetaLight) uses the negative queue length as a reward, which does not seem to be prone to this issue: sacrificing throughput would mean having longer queues, hence an optimal policy would not do that when one uses this reward function. For this reason, we keep the MetaLight original setup and evaluate the performance on the basis of Travel Time only.

\paragraph{Statistical Distance Metric (KL Distance)} In this paper, we will test models trained in a distribution $P_{train}$ on the test distribution(s) $P_{test}$. To measure how different the test distribution is from the train distribution, we use the Kullback-Liebler Divergence, which we call KL Distance ($D_{KL}$) from this point onwards. A traffic distribution is characterized by the portion of vehicles over the total in each traffic movement (see Figure \ref{fig:traffic_movements}). Letting $n(i)$ be the number of vehicles in traffic movement $i$, the distribution is therefore
\begin{equation}
    P(i) = \frac{n(i)}{\sum_{j}n(j)}
\end{equation}
Given $P_{train}$ and $P_{test}$, the KL Distance is then computed as
\begin{equation}
    D_{KL}(P_{train}|P_{test}) = \sum_{i} P_{train}(i) \log \frac{P_{train}(i)}{P_{test}(i)} 
\end{equation}

\section{Experiments}
We perform a wide range of experiments to evaluate MetaLight performance in a synthetic scenario, where we set the desired levels of traffic in the training set and the level of distributional shift in the test set. Then, we evaluate MetaLight on a realistic scenario, where we simulate the real-world distributional shift occurring in a US city as measured by on-site sensors. \hl{Finally, we perform an ablation study of some key MetaLight hyperparameters to validate our hypotheses about the failure cases.}

\subsection{Synthetic Dataset}
\label{sec:synthetic_dataset}
\subsubsection{Training set}
We generate a training set by first deciding on five base traffic distributions. Each traffic distribution is characterized by the number of vehicles that will flow through each of the eight possible movements (phases) in one hour of operation. From these base distributions, we generate 25 scenarios by:
\begin{itemize}
    \item Modifying each base distribution by increasing/reducing the volume of all traffic movements uniformly by -20\%, -10\%, 0\%, +10\%, +20\%
    \item Increasing/reducing individually the traffic volume of each movement by a random value $\pm20$\%
    \item Setting the arrival times of each vehicle by random sampling in the interval [0s, 3600s] (i.e., the hour of simulation into consideration) 
\end{itemize}
The resulting 25 scenarios constitute the training set for both MetaLight and the FRAP++ model. 

\subsubsection{Test scenarios}
\begin{figure}
    \centering
    \includegraphics[width=0.5\textwidth]{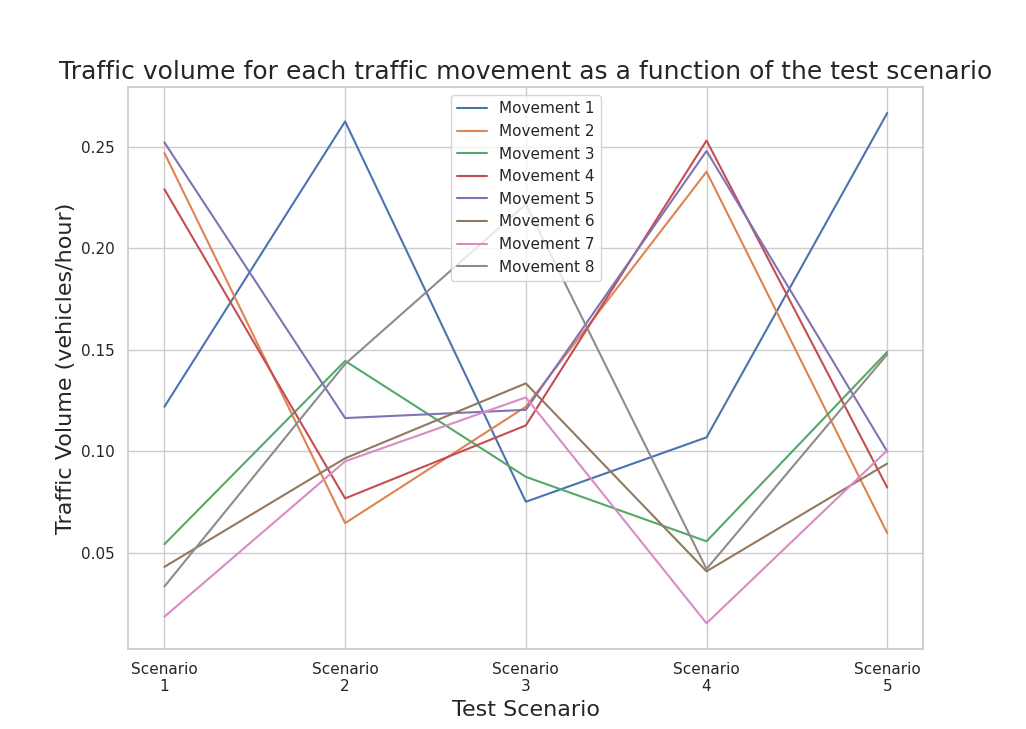}
    \caption{Traffic volume for each traffic movement (phase) across scenarios. On the X axis, we show the scenario number, and on the Y axis the traffic volume. Each line represents a traffic movement. The plot highlights the high variability in the traffic distribution since each traffic movement is subject to very different levels of traffic across scenarios.}
    \label{fig:moves_by_scenario}
\end{figure}

We generate 3 test scenarios following the same process described above, but increasing the variability of traffic flow by an additional 15\%, to generate traffic distributions that shift from the training ones. We generate 2 additional test scenarios where the relative traffic volume across movements is closer to the training set (max 10\% variation in step 1) but the total traffic volume is uniformly increased by 30\%. We can therefore observe how MetaLight performs when the relative volume across traffic movements shifts significantly from the training distribution and when the total volume of traffic is higher than the one observed during training. Table \ref{tab:distributions} shows the traffic volumes of each scenario and Figure \ref{fig:moves_by_scenario} shows a graphical view of the table to highlight the degree of variation in traffic volume for each traffic movement across test scenarios.

\begin{table*}
\caption{Traffic volume in vehicles per hour  (percent of total) for each traffic movement in the 5 base traffic distributions used to generate the synthetic training set}
\begin{center}
\begin{tabular}{|c|c|c|c|c|c|c|c|c|c|} 
 \hline
 Scenario & Mov. 1 & Mov. 2 & Mov. 3 & Mov. 4 & Mov. 5 & Mov. 6 & Mov. 7& Mov. 8 & Total volume \\
 \hline
1 & 98 & 159 & 114 & 147 & 157 & 174 & 165 & 289 & 1313\\ 
& (7.52\%) & (12.2\%) & (8.74\%) & (11.28\%) & (12.04\%) & (13.35\%) & (12.66\%) & (22.17\%) & (100\%)\\
\hline
2& 164 & 332 & 73 & 308 & 339 & 58 & 25 & 45 & 1344\\ 
& (12.2\%) & (24.7\%) & (5.43\%) & (22.91\%) & (25.22\%) & (4.31\%) & (1.86\%) & (3.34\%) & (100\%)\\
\hline
3& 345 & 85 & 190 & 101 & 153 & 127 & 125 & 188 & 1314\\ 
& (26.25\%) & (6.46\%) & (14.45\%) & (7.68\%) & (11.64\%) & (9.66\%) & (9.51\%) & (14.3\%) & (100\%)\\
\hline
4 & 188 & 418 & 98 & 445 & 436 & 72 & 27 & 74 & 1758\\ 
& (10.69\%) & (23.77\%) & (5.57\%) & (25.31\%) & (24.8\%) & (4.09\%) & (1.53\%) & (4.2\%) & (100\%)\\
\hline
5 & 451 & 101 & 252 & 139 & 169 & 159 & 170 & 250 & 1691\\ 
& (26.67\%) & (5.97\%) & (14.9\%) & (8.21\%) & (9.99\%) & (9.4\%) & (10.05\%) & (14.78\%) & (100\%)\\
\hline
\hline
\end{tabular}
\label{tab:distributions}
\end{center}
\end{table*}

\begin{figure}
    \centering
    \includegraphics[width=0.5\textwidth]{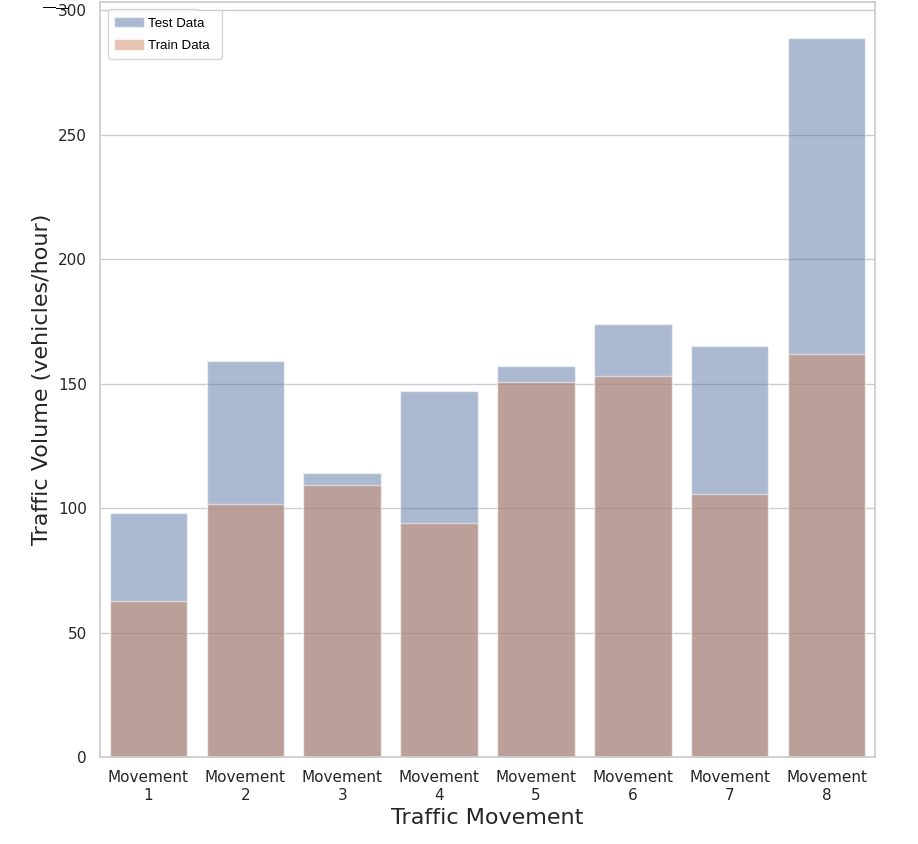}
    \caption{Comparison between Scenario 1 traffic distribution and the average training distribution. On the X axis the traffic movements from phases 1 to 8, on the Y axis the traffic volume (vehicles per hour). The corresponding KL Distance is 0.209}
    \label{fig:scenario_3}
\end{figure}

\subsection{Real World Distributional Shift}
\begin{figure}
    \centering
    \includegraphics[width=0.45\textwidth]{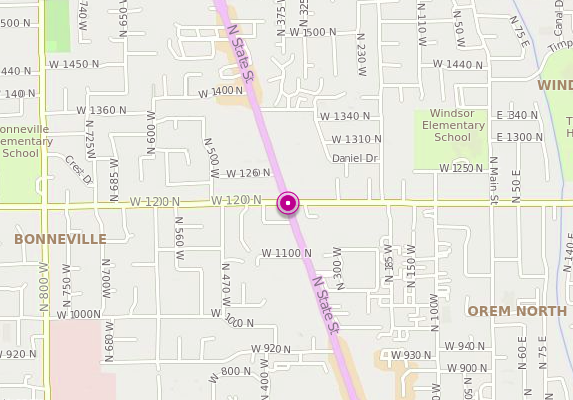}
    \caption{Intersection between US-89 and W 1200 N in Orem, Utah (US)}
    \label{fig:6394}
\end{figure}

In this experiment, we use real-world data for afternoon peak hours at an intersection in Salt Lake City, Utah. Train and test sets are generated as in the experiment above, but instead of manually selecting the levels of traffic, we use the real-world data in our possession. Data contains traffic volumes for every traffic movement at the intersection between US-89 and W 1200 N in Orem, Utah (US) in 5-minute time buckets. Figure \ref{fig:6394} shows a portion of the Orem map containing the intersection. This data is obtained from the Automated Traffic Signal Performance Measures (ATSPM) portal of the Utah Department of Transportation (UDOT). Our software pipeline allows us to easily retrieve data from ATSPM in a suitable format at any date and time for intersections in the whole state of Utah state where ATSPM is used.

%

\subsubsection{Training and Test Sets}
We generate three base traffic distributions from real traffic data in the AM peak (8-9 AM), Midday peak (2-3PM), and PM peak (5-6 PM) and, for each, we generate a training set as described in Section \ref{sec:synthetic_dataset}. We chose these 1-hour scenarios as they correspond to the three main hours of peak traffic and are those that most interest traffic signal engineers and DOTs. We generate three test scenarios, one for each of the three base distributions. We then performed two experiments, called Experiment 1 and Experiment 2. In Experiment 1, we use the AM peak dataset as the training data set, and we test on the Midday peak and PM peak test scenarios. In Experiment 2, we train on the Midday peak dataset and test on the AM peak and PM peak test scenarios.

\begin{table*}
\caption{Traffic volume in vehicles for each traffic movement for AM peak (8-9 AM) Midday peak (2-3 PM), and PM peak (5-6 PM) scenarios.}
\begin{center}
\begin{tabular}{|c|c|c|c|c|c|c|c|c|c|} 
 \hline
 Scenario & Mov. 1 & Mov. 2 & Mov. 3 & Mov. 4 & Mov. 5 & Mov. 6 & Mov. 7& Mov. 8 & Total volume \\
\hline
AM Peak& 45 & 218 & 58 & 290 & 30 & 476 & 54 & 65 & 1236 \\
\hline
Midday Peak& 36 & 101 & 35 & 309 & 53 & 415 & 49 & 288 & 1251 \\
\hline
PM Peak& 93 & 304 & 87 & 446 & 89 & 358 & 107 & 489 & 1973 \\
\hline
\hline
\end{tabular}
\label{tab:udot_shift}
\end{center}
\end{table*}

\subsection{Ablation Study: Gradient Steps}
\hl{In our Discussion section, we hypothesize that the number of meta-gradient steps should have an important impact on the RL model-free algorithm that MetaLight uses. Therefore, we perform an ablation study where we train several versions of the MetaLight paper model (on the same data as in the paper) for an increasing number of meta-gradient steps. What we hypothesize is that an increasing number of steps should improve performances, although it might be more difficult to train.

We therefore train several models for 1, 2, 3, 5, 10 gradient steps and perform the meta-adaptation as described in the MetaLight Section. We then test the model on the paper test scenarios, and observe how the travel time varies as a function of the number of gradient steps. }

\section{Results}
\subsection{Synthetic Dataset}
\begin{figure}
    \centering
    \includegraphics[width=\linewidth]{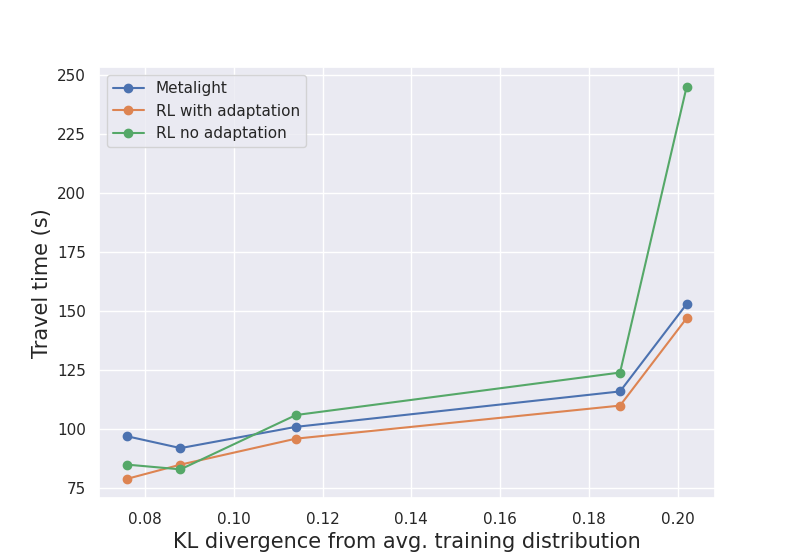}
    \caption{Travel time as a function of KL Distance from the training distribution for MetaLight, RL with adaptation, and RL with no adaptation.}
    \label{fig:synthetic_results_kl}
\end{figure}

Results for the 5 scenarios and each algorithm are shown in Table \ref{tab:results_travel_time}. The plot of Figure \ref{fig:synthetic_results_kl} depicts these results as a function of the distance between the test and training distributions, measured with the KL Distance metric, as described in Section \ref{sec:metrics}. Table \ref{tab:times} shows a comparison of the time it takes for MetaLight base model training, MetaLight adaptation step, and FRAP++ training.

We can observe how, in these synthetic traffic scenarios, re-training a Reinforcement Learning agent from scratch is superior to the MetaLight adaptation. MetaLight shows small drops in performance in Scenarios 3, 4, and 5 (+4\%, +4\%, and +5\% travel time), and two more severe drops in Scenarios 1 and 2 (+22\% and +12\%). The most severe (Scenario 1), corresponds to one of the three scenarios where there is a large variation in the proportion of traffic across movements compared to the training scenarios. The moderate performance drop (Scenario 2) also falls in this category, while in Scenario 3 as well as in the two scenarios of increasing traffic volume (Scenario 4 and 5) MetaLight successfully adapts with small performance degradations. In Figure \ref{fig:scenario_3} we show a comparison between the average training traffic distribution and the traffic distribution of Scenario 1, which is the one where MetaLight performs worse. 

For reference, in the third row of Table \ref{tab:results_travel_time} (RL with no adaptation) we also show the performance of the FRAP++ model trained for Scenario 1 when deployed on the other four scenarios. This is intended to provide insight into what happens if one naively applies to a new scenario a model trained for a particular traffic distribution. As expected, the performance is poor, ranging from a +7\% to a +66\% travel time increase.  

Figure \ref{fig:synthetic_results_kl} shows how the degradation in performance clearly correlates with an increase in KL Distance from the training distribution. This is true for both types of RL models (with and without adaptation), but also for MetaLight. Although for small values of KL divergence (up to 0.12) its performance remains at the same level, it suffers from substantial degradation at the higher end of the scale.

\begin{table*}
\caption{Average travel time (seconds) for synthetic traffic}
\begin{center}
\begin{tabular}{ |c|c|c|c|c|c|} 
 \hline
 Algorithm & Scenario 1 & Scenario 2 & Scenario 3 & Scenario 4 & Scenario 5\\
\hline
MetaLight & 97s (+22\%) & 96s (+12\%) & 101s (+5\%) & 116s (+5\%) & 153s (+4\%) \\
RL with adaptation& \textbf{79s} & 85s (+2.4\%) & \textbf{96s} & \textbf{110s} &\textbf{147s} \\
RL with no adaptation& 85s (+7\%) & \textbf{83s} & 106s (+10\%) & 124s (+12\%) &  245s (+66\%) \\
 \hline
\end{tabular}
 \label{tab:results_travel_time}
\end{center}
\end{table*}

\subsection{Real World Distributional Shift}
Tables \ref{tab:real_world_res_1} and \ref{tab:real_world_res_2} show the results for the real-world distribution shift scenarios (Experiment 1 and Experiment 2). While MetaLight performs correctly in the AM peak and Midday peak scenarios, showing some promising degree of adaptation, it does not adapt well to the PM peak scenario, which is the most difficult one as the distribution is very different from the other two. In both cases, RL with no adaptation outperforms MetaLight and the naive RL adaptation method. We highlight in boldface the best values for each scenario.

\begin{table*}
\caption{Travel time (seconds) for traffic from real-world data (Experiment 1)}
\begin{center}
\begin{tabular}{ |c|c|c|c|} 
 \hline
 Algorithm & Scenario 1 & Scenario 2 & Scenario 3 \\
\hline
MetaLight & \textbf{75s}& \textbf{68s} & 96s \\
RL with adaptation & 76s & 82s & \textbf{87s}\\
RL with no adaptation & 79s & 71s & 102s\\
 \hline
\end{tabular}
 \label{tab:real_world_res_1}
\end{center}
\end{table*}

\begin{table*}
\caption{Travel time (seconds) for traffic from real-world data (Experiment 2)}
\begin{center}
\begin{tabular}{ |c|c|c|c|} 
 \hline
 Algorithm & Scenario 1 & Scenario 2 & Scenario 3 \\
\hline
MetaLight & \textbf{61s} & 63s & 90s \\
RL with adaptation& 65s & \textbf{62s} & 89s\\
RL with no adaptation& 77s & 71s & \textbf{85s}\\
 \hline
\end{tabular}
 \label{tab:real_world_res_2}
\end{center}
\end{table*}

\subsection{Ablation Study: Meta-Gradient Steps}
\begin{figure}
    \centering
    \includegraphics[width=0.5\textwidth]{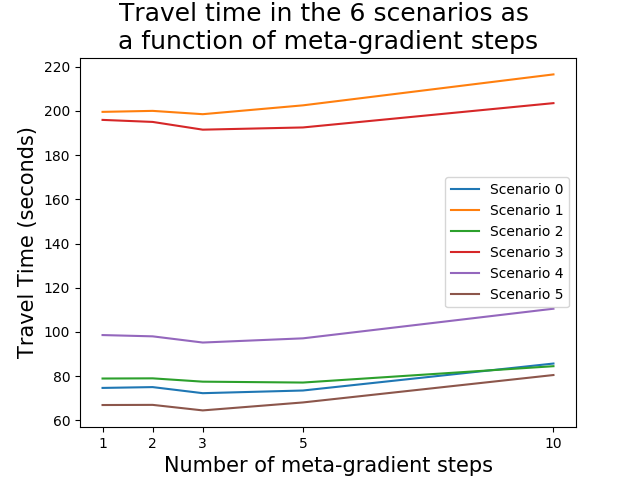}
    \caption{\hl{Travel time as a function of the number of meta-gradient steps during adaptation, data recorded in the 6 
 scenarios of the MetaLight paper.}}
    \label{fig:meta_grad}
\end{figure}

\hl{Figure} \ref{fig:meta_grad} \hl{shows the results for this ablation study. Results show that an increase in meta-gradient steps indeed makes MetaLight perform better, but only for a small amount, as travel time starts growing again after 3 gradient steps. In Section} \ref{sec:discussion_causes}, \hl{we discuss the consequences of this outcome.}

\subsection{Training and adaptation times}
Looking at the times in Table \ref{tab:times} we see that although the MetaLight base model training takes some 25\% more time than just training a FRAP++ model, the adaptation step is around 120 times faster than training a new FRAP++ model from scratch. Clearly, this is very significant.

\begin{table}
\caption{Time required for each task}
\begin{center}
\begin{tabular}{|c|c|} 
 \hline
 Task & Time \\
\hline
MetaLight training base model & $\approx$ 2.5h \\
MetaLight adapting base model & $\approx$ 2min\\
FRAP++ training & $\approx$ 2h\\
\hline
\end{tabular}
\label{tab:times}
\end{center}
\end{table}

\section{Discussion}
In some cases, the results shown above indicate that MetaLight can be a good option when it is deployed for new scenarios. Despite some loss in performance, it allows us to obtain a Reinforcement Learning model ready to operate in the new scenario in just a matter of minutes. When scaling to more complex models and multi-agent settings (not explored in this paper), where training times increase dramatically, the MetaLight approach could save even days or weeks of training.

However, in our experiments, we also observed many instances in which this adaptation step did not allow MetaLight to properly handle traffic distributions significantly different from the training ones. In particular, MetaLight does not manage to adapt to test scenarios that are too different from those observed in training, \hl{as per our KL distance measure.} As shown in the Results section, an increase in the distance of the test data distribution from the training distribution (measured as KL Distance) is generally related to performance degradation for all models under observation, including MetaLight. While MetaLight's performance remains stable for low levels of KL Distance, it suffers significant degradation at higher levels. This poses some limitations and challenges on the possibility of having fully RL-controlled traffic lights, as large distributional shifts do happen in the real world. \hl{Moreover, although Metalight represents an important first step toward building foundational models for traffic signal control, it still exhibits failure cases that highlight the limitations of this approach in handling the distribution shift problem.}



\subsection{Possible causes of sub-optimal performance}
\label{sec:discussion_causes}
\hl{Deep RL agents exploit the \textit{bootstrapping} technique to learn optimal policies. As described in Section} \ref{sec:rl}, \hl{when learning the Q function for state-action pair} $s, a$ \hl{we use the current estimate of} $Q(s', a')$. \hl{This not only allows us to learn Q values faster (as opposed to Monte Carlo methods) but has some interesting implications on how information about the reward of a particular state "backpropagates" through previous states and Q value estimates.}

\begin{figure}
    \centering
    \includegraphics[width=\linewidth]{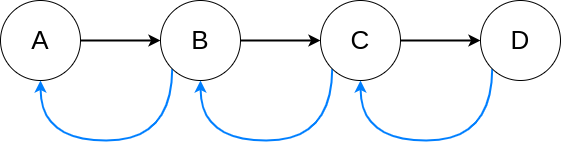}
    \caption{Bootstrapping in RL. The agent moves forward in time, from state A to state D. Rewards propagate in the opposite direction, i.e. only when the reward for D is known the agent can estimate the reward-to-go of C, which leads to an estimate of that of B, and finally of A.}
    \label{fig:bootstrapping}
\end{figure}

\hl{Imagine we have four states A, B, C, and D (see Figure }\ref{fig:bootstrapping}). \hl{We observe the set of transitions} $\{(A \rightarrow B), (B \rightarrow C), (C \rightarrow D)\}$ \hl{and the corresponding rewards} $r_{AB}, r_{BC}, r_{CD}$. \hl{To learn the Q value of the action that leads us to state} $B$ \hl{starting from state} $A$ \hl{we need first to know the Q values of the other transitions. Equation} \ref{eq:bellman_loss} \hl{uses the }\textit{current} \hl{estimate of the Q value of the \textit{following} state (i.e., }$Q(s', a')$ \hl{to allow us to improve our estimate of the preceding state. For the first transition, that would correspond to minimizing} $\left(Q(A, a) - (r_{AB} + \gamma \max_{a'}Q(B, a'))\right)^2$ and, if $Q(B, .)$ \hl{was not updated in a previous iteration, we would be using a wrong target. It is therefore essential to minimize the loss of Equation} \ref{eq:bellman_loss} \hl{in multiple iterations when learning from batches of examples.}

\hl{In conventional RL training, the number of these iterations is very high, and we are generally confident to converge to the optimal} $Q$ \hl{value. In the MetaLight adaptation step, however, the number of iterations is low (e.g., 1-5) and we do not have any guarantee that the "backpropagation" of rewards is sufficient. This is especially true for scenarios where the optimal Q-value function is very different from that of training scenarios, and would therefore require more adaptation steps. A reason for the failures reported in the results could be that the adaptation step of MetaLight is not sufficient to correctly capture the reward backpropagation process --typical of RL training-- and when the distribution shifts significantly the approximation error of the $Q$ function we obtain becomes too large.

Our experiments on meta-gradient steps seem to partially confirm this since we achieve some performance improvement as we increase the number of adaptation steps. However, as the number of gradient steps surpasses a low value (around 4), MetaLight performance begins degrading. This could mean that it is too difficult to meta-train a model to be updated with a high enough number of updates to make it suitable for model-free adaptation.}

\subsection{Areas of Improvement}
\hl{We believe that further improvement of the MetaLight framework is possible in several areas, from data-generation to algorithmic refinements.} In \cite{generalight}, \hl{a generative-adversarial approach is used to augment the training data used in the MetaLight training. However, this approach is only used to generate larger datasets that are similar to the existing ones, providing limited samples on possible scenarios with higher levels of distributional shift. We propose to improve this approach by modifying the adversarial component to include the agent performance in its cost function, pushing it to generate scenarios that are more and more difficult for MetaLight to adapt to. On the algorithmic side, we believe that performing a larger number of adaptation steps could improve performance. However, such steps likely need to be performed carefully, with smaller learning rates or techniques to minimize policy drift (such as trust regions or clipped objectives). In fact, our experiments show that although an increase in adaptation steps seem to initially improve performance, it quickly "destroys" the agent's policy, leading to a worsening in performance. Finding a method to perform a larger number of adaptation steps without incurring overfitting or over-propagation of errors could be the key to expand MetaLight's adaptation capabilities. Further research is required to explore this possibility.}

\subsection{Next Steps in Distributional Shift Adaptation}
The workflow represented by MetaLight consists of a) training an adaptable model; b) collecting data from a new scenario in which one wants to deploy it; and c) adapting the model to that particular scenario. This mechanism can be useful to reduce the deployment time of Deep RL models on new intersections but does not address a central issue in the real-world deployment of Deep Learning models for Traffic Signal Control; i.e., the variation of traffic distribution after the model has been deployed. Even when a large amount of data has been collected and the model has been deployed, the intersections' traffic patterns can still change unpredictably. This can happen due to unexpected events, such as road closures, accidents, and the intrinsic non-stationarity of traffic patterns. In fact, in ever-evolving cities, traffic patterns vary over time as the city changes, new places of interest are built or moved, means of transportation change, and public transport improves.

Deep RL models for Traffic Signal Control will therefore need to be robust to these changes to be safely deployed in our cities. Robustness can be achieved either through: a) the ability to monitor and detect that the observed traffic distribution is beyond the reliable regime of operation of the deep model and a different type of policy needs to be used; or b) the development of deep models that are able to safely deal with distribution shift scenarios with low degradation in performance.

\section{Conclusions}
In this paper, we investigated a state-of-the-art Meta-Learning algorithm designed to deal with the Distributional Shift problem in AI-based Transportation Networks and Traffic Signal Control by exploiting the Meta-Learning framework. Despite some encouraging results, we observe instances in which such an approach does not perform well. 

This paper should therefore be viewed as a cautionary note for practitioners in Reinforcement Learning and Traffic Signal Control and their real-world deployment: our results suggest that the current state of research is not mature enough to directly handle all the distribution shift scenarios with Reinforcement Learning and Meta-Learning techniques. 

\hl{While Metalight can be viewed as a step forward in building foundational models for traffic signal control, our study demonstrates that the distributional shift problem remains as a major concern for reliable deployment of deep reinforcement learning models in the real world. Foundational models (large models trained on a larger amount of data) are currently a hot topic in machine learning and large language models (LLM); however, our experiments demonstrate that foundational models might not always be able to solve the distribution shift (or domain adaptation) problem in machine learning.}

Further research is needed to fully understand the root cause of why in certain cases such approaches using Model-Agnostic Meta-Learning do not lead to good or even acceptable results.

\end{document}